# IEEE Copyright Notice

# Automated Knowledge Modeling for Cancer Clinical Practice Guidelines

Pralaypati Ta, Bhumika Gupta, Arihant Jain, Sneha Sree C, Arunima Sarkar, Keerthi Ram, Mohanasankar Sivaprakasam

*Abstract*— Clinical Practice Guidelines (CPGs) for cancer diseases evolve rapidly due to new evidence generated by active research. Currently, CPGs are primarily published in a document format that is ill-suited for managing this developing knowledge. A knowledge model of the guidelines document suitable for programmatic interaction is required. This work proposes an automated method for extraction of knowledge from National Comprehensive Cancer Network (NCCN) CPGs in Oncology and generating a structured model containing the retrieved knowledge. The proposed method was tested using two versions of NCCN Non-Small Cell Lung Cancer (NSCLC) CPG to demonstrate the effectiveness in faithful extraction and modeling of knowledge. Three enrichment strategies using Cancer staging information, Unified Medical Language System (UMLS) Metathesaurus & National Cancer Institute thesaurus (NCIt) concepts, and Node classification are also presented to enhance the model towards enabling programmatic traversal and querying of cancer care guidelines. The Node classification was performed using a Support Vector Machine (SVM) model, achieving a classification accuracy of 0.81 with 10-fold cross-validation.

## I. INTRODUCTION

Knowledge about cancer therapies and treatment protocols is rapidly evolving, owing to the continued inflow of findings and new evidence being generated from clinical trials. Maintaining this evolving knowledge with naive documentation and version control has led to regularly updated clinical practice guidelines (CPGs). National Comprehensive Cancer Network (NCCN) regularly publishes CPGs [1] containing treatment recommendations for multiple cancers. These guideline documents attempt to be both easy to consume and reference by caregivers and reflect up-to-date information and evidence. Simple documentation and management methods could rapidly become untenable, in the face of the evolving nature of knowledge. The above needs are better addressed by machine-readable knowledge bases [2]. The content of the CPG documents has been majorly graphical, for ease of consumption. In knowledge modeling for CPG, interoperability, and conversion among common formats (like flow diagrams) are essential, as a means of verifying the knowledge model and supporting prevalent formats. Ontology-based modeling has been proposed in the literature [3], [4], [5]. Ontologies are developed to convert unstructured information from the guidelines into structured knowledge [5].

There is a need to develop an automated method for parsing CPG documents to generate knowledge objects and linkages that authentically capture graphical cancer care pathway information in the NCCN CPGs and express the knowledge in a programmatically traversable and queryable fashion.

Qiu et al. [6] proposed a framework for the automated encoding of NCCN guidelines into computer-interpretable guidelines (CIGs). The NCCN CPGs are published in PDF format. Their work converted PDF pages to image format first and then image processing techniques were used to convert them to semi-structured guidelines. Liu et al. [7] proposed a three-layer knowledge unit representation model for NCCN guidelines. The NAVIFY [8] digital solution converts the NCCN guideline into a tree and step view to improve the usability of the guidelines. The proposed work, however, directly captures the graphical treatment pathway from the guideline documents. It parses the PDF object model to extract the knowledge items and their relationships.

Unstructured medical text is not amenable to programmatic interaction. The mapping of unstructured medical texts to standard concepts, such as UMLS Metathesaurus [9] and NCI Thesaurus (NCIt) [10], has been proposed in the literature to generate a structured data model. Bissoyi et al. [11] proposed an algorithm to extract UMLS concepts from medical discharge summaries. Schafer et al. [12] used UMLS mapping to generate word embedding for discharge summaries. Tao et al. [13] argued that NCIt mapping facilitates the utilization of the cancer registry data and implemented a mapping framework for the same.

The main contributions of this work are as follows:

- We propose an algorithm for automated retrieval of the knowledge components from NCCN guidelines. The algorithm retrieves the guideline knowledge by parsing the PDF documents and extracting the native PDF objects.

- We also propose a suitable data model to define and store the knowledge contained in these guidelines. We provide a schema to represent the retrieved guideline knowledge in JSON-LD [14] format.

- To make our knowledge model programmatically queryable, three enrichment techniques have been proposed in this work. 1) Enrichment with cancer

Pralaypati Ta, Bhumika Gupta, Arihant Jain, Sneha Sree C, Arunima Sarkar, Keerthi Ram and Mohanasankar Sivaprakasam are with the Healthcare Technology Innovation Centre (HTIC), Indian Institute of Technology Madras, India (e-mail: pralaypati@htic.iitm.ac.in).

Pralaypati Ta, Bhumika Gupta, Arihant Jain, Sneha Sree C, Arunima Sarkar and Mohanasankar Sivaprakasam are also with the Department of Electrical Engineering, Indian Institute of Technology Madras, India.

staging information by extracting them from the unstructured guideline content. 2) Mapping of the unstructured text to UMLS Metathesaurus [9] and NCI Thesaurus (NCIt) [10] concepts. 3) Classifying the text blocks into one of the five classes i.e. *Evaluation*, *Result*, *Decision*, *Action*, and *Uncertain*.

We evaluated the knowledge extraction algorithm with two different versions of NCCN treatment guideline document for Non-Small Cell Lung Cancer (NSCLC) to show that our algorithm can extract the guideline knowledge from both versions with high accuracy.

## II. METHODOLOGY

The NCCN guideline documents contain various content elements such as text blocks (called nodes), unordered and ordered lists, footnotes, links, tables, etc. to specify the treatment recommendations. The text blocks are connected by arrows and hyperlinks to represent the sequential ordering of the treatment pathway. The text blocks are also placed under different headings, called labels, to categorize the recommendations.

The PDF file format [15] lacks the concept of text lines, paragraphs, or arrows; various heuristics based algorithms have been developed to detect these high level objects. Fig. 1 shows the processing steps of knowledge model generation.

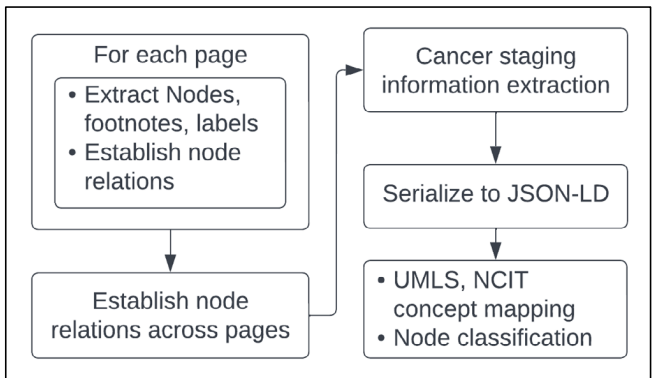

Figure 1. Knowledge model generation steps.

## III. SYSTEM IMPLEMENTATION

### A. Knowledge extraction & serialization

Apache PDFBox [16] library, version 2.0.25, was used to parse and extract the PDF objects from the guideline documents. PDFBox offers a capability in which the individual characters are grouped into text lines using character positions. An algorithm was developed building upon that capability for the extraction of the other high-level objects.

The nodes are calculated by grouping the content lines vertically close to each other. The algorithm also considers the vertical lines around the text blocks while generating the blocks to take care of the scenarios where the text lines are close enough but they are not part of the same block. Fig. 2 shows guideline text blocks along with the bounding box of the text blocks detected by this algorithm. As we can see, the text lines for different blocks are close enough, but the vertical lines around the block indicate that they are part of separate nodes.

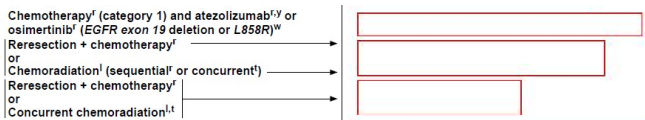

Figure 2. Detected text blocks, also called Nodes.

The relation between the nodes is represented using a directed arrow in the guideline document. The arrows are drawn using lines with tiny filled triangles at one end. To extract the relationships, the following steps were used.

- Extract the lines and the triangles.
- Find the association between the line and triangles to establish the source and target end of the lines.
- Find the text blocks close to the source and target end of a line.
- Establish the 'previous' and 'next' relationship between these two blocks.
- For cross-page links, establish the 'previous' and 'next' relationship between the last nodes of a page and the first nodes of the target page.

Fig. 3 shows a guideline page along with the extracted text blocks and the relationships between them. To extract tabular data, the open-source tool, called Tabula [17] has been used. Tabula offers several output formats for the data, such as JSON, CSV, and TSV. The JSON format was used so that it can be easily stored in the JSON-LD file.

We propose a JSON-LD schema for representing the guideline knowledge. The schema has elements for representing the treatment recommendation nodes along with their attributes, footnotes, and labels. The schema also includes elements for representing the relation between the nodes and the footnote references.

### B. Enrichment

*1) Extraction of Cancer Stages and T, N, M scores*

Regular expressions have been used to extract the cancer stages and T, N, and M scores from the content of the guideline nodes [18]. Different regular expressions have been designed for cancer stages and the T, N, and M scores. Cancer stages have the values specified in roman numerals I, II, III, or IV followed by an optional single A, B, or C character. N scores have the first character as N followed by a single digit number between 0 to 3. M scores have the first character M followed by a single digit number 0 or 1, followed by an optional single a, b, or c character. T scores have a similar pattern, first character T followed by a single digit number between 0 to 4, followed by an optional single a, b, or c character.

*2) UMLS and NCIT Concept Mapping*

ScispaCy [19] was used for UMLS and NCIT concept mapping. The biomedical entities were extracted from the content of the guideline text nodes using the en_core_sci_sm model of scispaCy. To improve the accuracy of downstream concept mapping, a few abbreviations such as CT, MRI, and RT have been expanded before passing the text to scispaCy. ScispaCy's UMLS entity linker module was used to link the extracted entities to the UMLS concepts. It performs char-3gram matching [19] of the entities with the UMLS concepts and returns multiple linked UMLS concepts, each with a score, for a single entity. The concept with the highest score, which is the first concept in the returned list, is used as a mapped concept for that entity. Also, to reduce the mapping errors for certain generic terms such as T3, T4, IIIA, etc., a custom mapping dictionary was used to map the concepts, instead of using the UMLS entity linker.

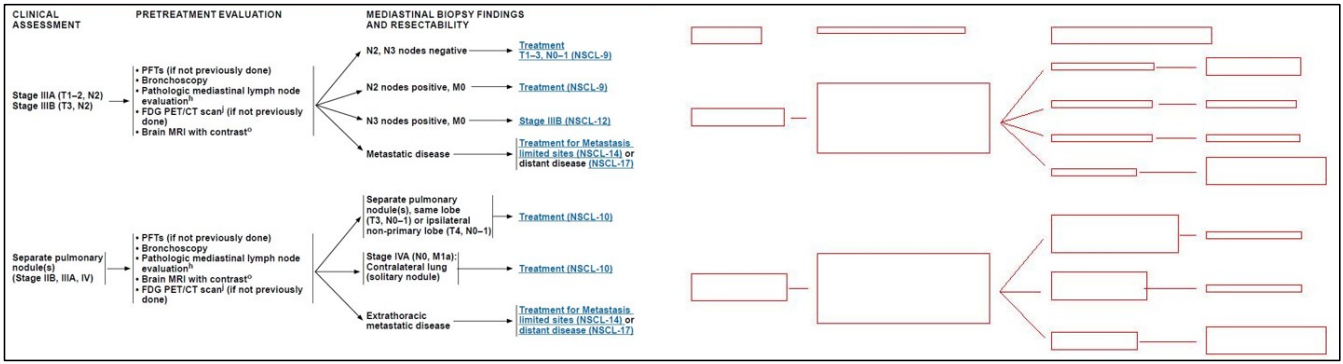

Figure 3. Extracted geometry of guideline treatment recommendation flow.

NCIT EVS search endpoint API [20] was used to map the extracted entities to NCIT concepts. The API provides various search options, such as contains, match, startsWith, phrase, etc. The first concept returned by the "contains" options was found to be the most accurate and was used as a mapped concept.

*3) Node Classification*

Node classification helps to understand the context of a particular node content. Nodes containing unique texts were identified first, then those are manually classified into the five classes, described below, to create the training dataset. An SVM model was trained using this dataset. The generated model is used to classify the nodes of new guideline versions.

- *Evaluation*: Nodes that indicate the clinical evaluations to be done, such as laboratory tests, scans, etc.
- *Result*: Nodes that contain information about the outcome of clinical evaluation, e.g. "Stable", "Progression" etc.
- *Decision*: Critical nodes which require an input/decision from the medical practitioner over an outcome of a particular sequence of action or as an outcome of a lab report/scan. Example: "Operable", "High risk" etc.
- *Action*: Nodes conveying information about the medical procedures/treatment to be done or some medication. Example: "CT at 6-12 mo", "No routine follow-up", "Durvalumab" etc.
- *Uncertain*: Nodes that contain other types of texts, e.g. nodes containing only hyperlinks to other nodes.

Features were extracted using bag-of-words representation and tf-idf weights. The classifier was built using Scikit Learn [21] library. A processing pipeline was created using CountVectorizer, TfidfTransformer, and SGDClassifier with hinge loss. GridSearchCV was used to perform 5-fold and 10-fold cross-validation using the StratifiedKFold strategy.

IV. RESULTS

*A. Knowledge Extraction*

Table I shows the result of the extraction algorithm. In version 1.2022 of the NCCN NSCLC guideline, there is a total of 914 nodes in the guideline treatment recommendation flow. The extracted JSON-LD file was imported into Neo4j [22] graph database to visually validate the treatment pathway. It can be observed that the extraction algorithm can extract the content with very high accuracy, resulting in a faithful representation of the guideline knowledge.

The extraction algorithm was run on the new version, 3.2022, of the NSCLC guideline to check if the algorithm is generic enough to handle the updated versions. The validation result shows that one additional connection error has been introduced for the new version.

*B. UMLS and NCIT Concept Mapping*

Table II shows the result of the concept mapping steps. The incorrect mappings are mainly due to sub-string based matching. The errors are high for NCIT mapping as the first result in the NCIT API response is not always the best matching concept.

*C. Node Classification*

In version 1.2022 of the NSCLC guideline, there are 439 unique nodes, out of 914 guideline nodes. Table III shows the best parameters reported by GridSearchCV for 5-fold and 10-fold CV. With these best parameters, the reported best score is 0.77 for a 5-fold CV and 0.81 for a 10-fold CV.

TABLE I. EVALUATION RESULT OF KNOWLEDGE EXTRACTION ALGORITHM

| Description | Count |
|---|---|
| Total no. of treatment recommendation nodes | 914 |
| Total no. of footnotes | 110 |
| Error in node formation | 7 |
| Error in connection detection between the nodes | 5 |
| Error in label assignment of nodes | 2 |
| Error in footnote detection and reference | 0 |

TABLE II. EVALUATION RESULT OF UMLS & NCIT CONCEPT MAPPING

| Concept Name | Total mapped entities count | Incorrect mapping count |
|---|---|---|
| UMLS | 1065 | 9 |
| NCIT | 1155 | 58 |

TABLE III. CROSS VALIDATION PARAMETERS

| Component | Parameter Name | Best Values (5 Fold) | Best Values (10 Fold) |
|---|---|---|---|
| CountVectorizer | max_df | 0.75 | 1.0 |
| | max_features | 5000 | 50000 |
| | ngram_range | (1, 2) | (1, 2) |
| TfidfTransformer | use_idf | False | True |
| | norm | L1 | L2 |
| SGDClassifier | max_iter | 50 | 10 |
| | alpha | 0.00001 | 0.00001 |
| | penalty | elasticnet | elasticnet |

## V. CONCLUSION

The proposed work describes an automated method for converting NCCN CPGs into a knowledge base. The method was validated using two versions of the NCCN Non-Small Cell Lung Cancer (NSCLC) treatment guideline and found that the generated knowledge base faithfully represents the guideline knowledge. The future scope of this work would be the enhancement and validation of the proposed method for the NCCN guidelines for other cancer types. Additionally, new enrichment techniques need to be developed to improve the query capability of the knowledge model.

## ACKNOWLEDGMENT

We thank Ravi Bhardwaj, Dr. Sanand Sasidharan, and Anuradha Kanamarlapudi for their valuable suggestions regarding the solution's architecture, technical development, and modeling aspects.